\documentclass[10pt,twocolumn,letterpaper]{article}

\usepackage[pagenumbers]{cvpr} %

\usepackage[dvipsnames]{xcolor}

\usepackage{multirow}
\usepackage{amsmath}
\usepackage{amsthm}
\usepackage[makeroom]{cancel}

\usepackage{makecell}
\usepackage{indentfirst}
\usepackage[dvipsnames]{xcolor}

\usepackage{booktabs,arydshln}
\usepackage{amssymb}%
\usepackage{pifont}%

\makeatletter
\def\adl@drawiv#1#2#3{%
        \hskip.5\tabcolsep
        \xleaders#3{#2.5\@tempdimb #1{1}#2.5\@tempdimb}%
                #2\z@ plus1fil minus1fil\relax
        \hskip.5\tabcolsep}
\newcommand{\cdashlinelr}[1]{%
  \noalign{\vskip\aboverulesep
           \global\let\@dashdrawstore\adl@draw
           \global\let\adl@draw\adl@drawiv}
  \cdashline{#1}
  \noalign{\global\let\adl@draw\@dashdrawstore
           \vskip\belowrulesep}}
\makeatother
\def\RA{\rlap{\scalebox{1.3}{\;$\rightarrow$}}}

\usepackage[accsupp]{axessibility} 

\def\RA{\rlap{\scalebox{1.3}{\;$\rightarrow$}}}

\definecolor{cvprblue}{rgb}{0.21,0.49,0.74}
\usepackage[pagebackref,breaklinks,colorlinks,allcolors=cvprblue]{hyperref}
\usepackage{multirow}

\def\model{\texttt{LoRACLR}}\title{\model{}: Contrastive Adaptation for Customization of Diffusion Models}

\author{\vspace{1mm}
    Enis Simsar$^{1}$
    \hspace{0.5cm}
    Thomas Hofmann$^{1}$
    \hspace{0.5cm}
    Federico Tombari$^{2,3}$
    \hspace{0.5cm}
    Pinar Yanardag$^{4}$
    \\
    $^1$ETH Zürich
    \hspace{0.25cm}
    $^2$TU Munich
    \hspace{0.25cm}
    $^3$Google
    \hspace{0.25cm}
    $^4$Virginia Tech\\
    {\tt\small \href{https://loraclr.github.io}{https://loraclr.github.io}}
}

\begin{document}

\twocolumn[{
\maketitle
\begin{center}
    \captionsetup{type=figure}
    \vspace{-1em}
\newcommand{\imwidth}{1\textwidth}

\begin{tabular}{@{}c@{}}
 
\parbox{\imwidth}{\includegraphics[width=\imwidth, ]{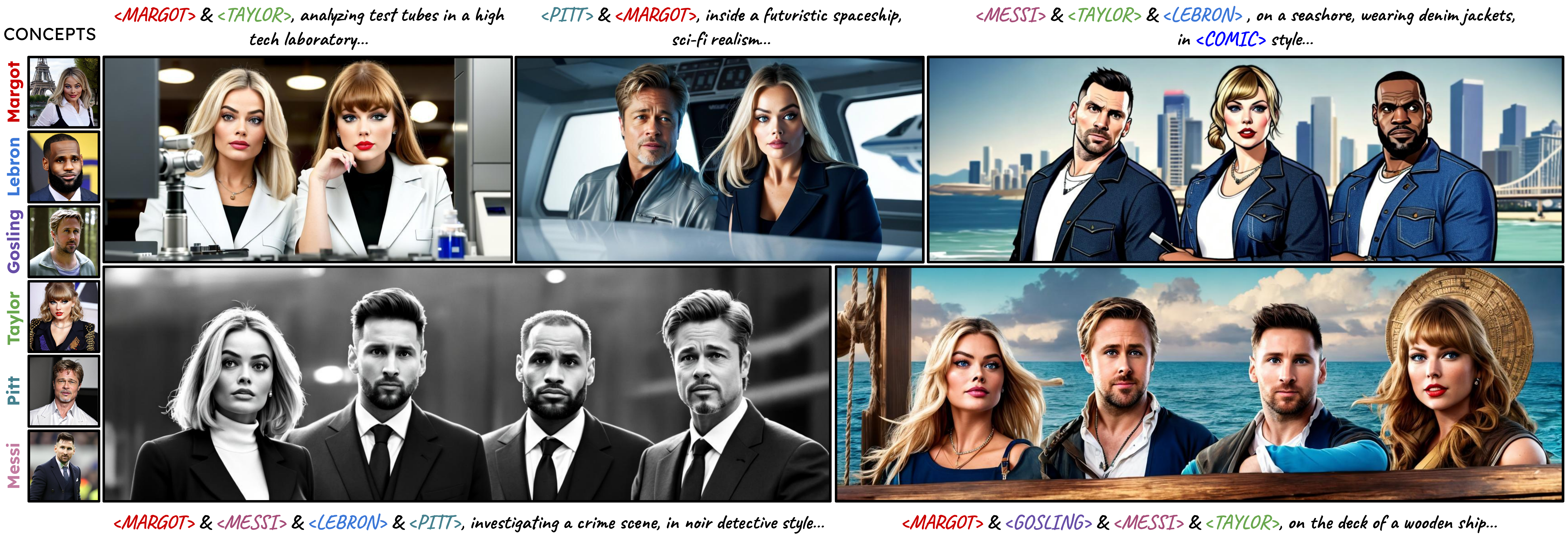}}
\\

\vspace{1em}
\end{tabular}
    \vspace{-2.5em}
    \captionof{figure}{\textbf{High-Fidelity Multi-Concept Image Generation.} Examples illustrating \model{}'s ability to generate unified scenes with multiple distinct characters and styles across diverse settings. Each scene demonstrates \model{}'s capability to combine varied concepts seamlessly, preserving the original identities of each character, as seen in concepts.}
 
    \label{fig:teaser}
    \vspace{0.5em}
\end{center}
}]

\maketitle
\begin{abstract}
Recent advances in text-to-image customization have enabled high-fidelity, context-rich generation of personalized images, allowing specific concepts to appear in a variety of scenarios. However, current methods struggle with combining multiple personalized models, often leading to attribute entanglement or requiring separate training to preserve concept distinctiveness. We present \model{}, a novel approach for multi-concept image generation that merges multiple LoRA models, each fine-tuned for a distinct concept, into a single, unified model without additional individual fine-tuning. \model{} uses a contrastive objective to align and merge the weight spaces of these models, ensuring compatibility while minimizing interference. By enforcing distinct yet cohesive representations for each concept, \model{} enables efficient, scalable model composition for high-quality, multi-concept image synthesis. Our results highlight the effectiveness of \model{} in accurately merging multiple concepts, advancing the capabilities of personalized image generation.
\end{abstract}
    
\section{Introduction}
\label{sec:intro}

Diffusion models for text-to-image generation~\cite{ddpm} have revolutionized image synthesis from textual prompts, as demonstrated by major advancements forward from Stable Diffusion~\cite{rombach2022high}, Imagen~\cite{imagen}, and DALL-E 2~\cite{dalle2}. Customization techniques for these models have further amplified their versatility, enabling the personalized image generation of specific concepts such as characters, objects, or artistic styles. Low-Rank Adaptation (LoRA)~\cite{hu2021lora} has emerged as a powerful tool for customizing pre-trained models with minimal retraining, allowing for flexible, efficient personalization. By combining LoRA with advanced customization methods like DreamBooth~\cite{ruiz2023dreambooth}, users can generate images that not only retain high-fidelity but also capture their unique creative vision.

However, combining multiple LoRA models to create a single composition remains a significant challenge. Current multi-concept models often struggle to maintain the quality of individual concepts, require simultaneous training on multiple concepts \cite{kumari2023multi}, or need per-image optimization \cite{meral2024clora}. Alternative methods face specific limitations: some can only merge style and content LoRAs~\cite{shah2023ziplora}, while others become unstable as the number of combined LoRAs grows~\cite{huang2023lorahub}. Methods like Mix-of-Show~\cite{gu2023mix} require specialized LoRA variants, such as Embedding-Decomposed LoRAs (EDLoRAs), which diverge from the standard LoRA formats widely used in the community. More recent approaches, like OMG~\cite{kong2024omg}, employ segmentation methods to isolate subjects during generation, but these approaches rely heavily on the accuracy of the segmentation models used. These challenges limit the broader applicability of text-to-image models, particularly when multiple distinct concepts must coexist in a single output image.

To address these challenges, we propose \model{}, a novel approach that combines multiple LoRA models into a single model capable of accurately generating multiple concepts simultaneously. Our method introduces a novel contrastive objective that aligns each model's weight space, prevents interference, and preserves fidelity by ensuring that each model represents its respective concept distinctly within the joint composition. Importantly, our approach allows for the use of pre-existing LoRA models without the need for retraining or accessing original training data. By employing contrastive learning, \model{} achieves scalable model composition, enabling high-quality multi-concept image generation without requiring additional fine-tuning or computational overhead. 

Our extensive evaluations reveal that \model{} achieves significant improvements in both visual quality and compositional coherence over baseline methods. Through qualitative and quantitative experiments, we demonstrate that our approach consistently preserves the fidelity and identity of each concept while avoiding the common issue of feature interference, even as complexity increases with additional concepts. \model{} provides a practical and scalable solution to compositional customization in generative models, with promising implications for applications like virtual content creation, personalized storytelling, and digital art.

\begin{figure*}[!htb]
\centering
\includegraphics[width=\textwidth]{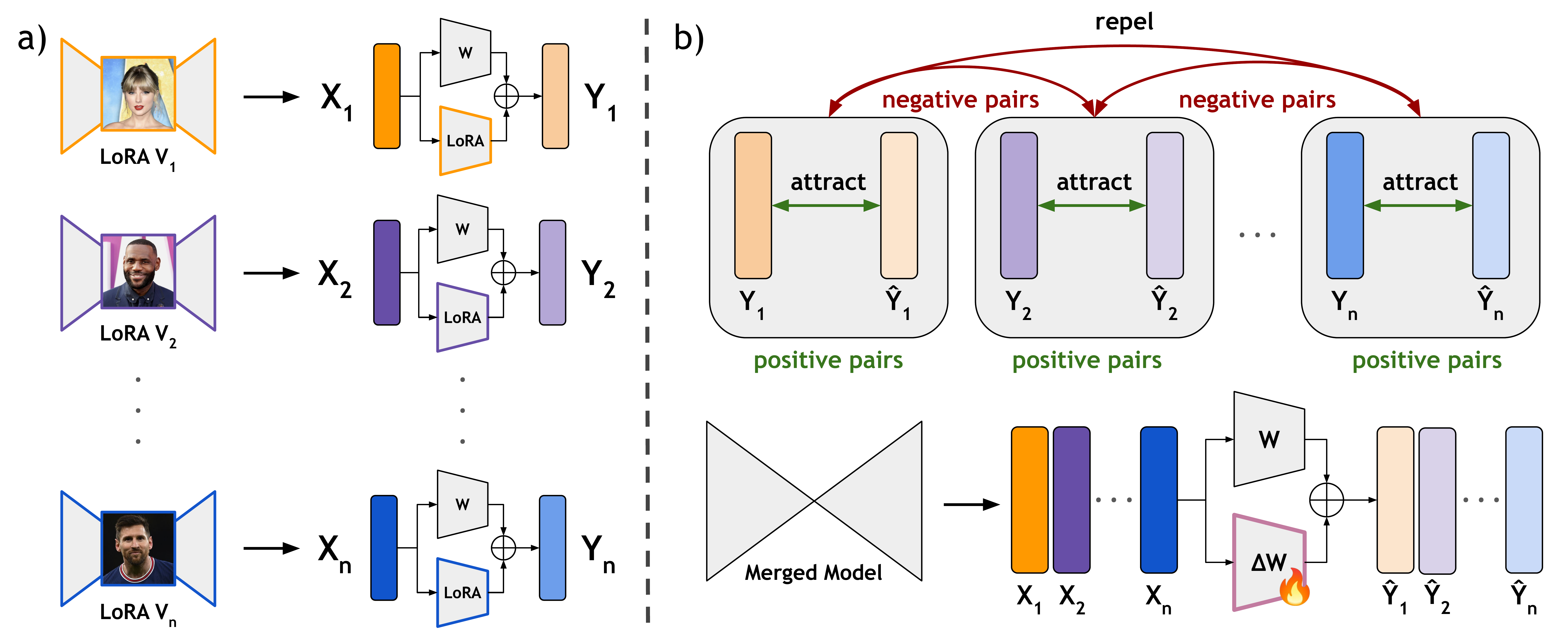}
    \vspace{-1em}
    \caption{\textbf{Framework Overview.} The framework comprises two main stages: (a) generating concept-specific representations with individual pre-trained LoRA models and (b) merging these representations into a unified model using a novel contrastive objective. In (a), each LoRA model produces input-output pairs $(X_i, Y_i)$ for distinct concepts $(V_1, V_2, \dots, V_n)$, establishing positive pairs (aligned concepts) and negative pairs (unrelated concepts). In (b), these representations are combined into a single model, $\Delta W$, to enable multi-concept synthesis. \model{} aligns attracting positive pairs to ensure identity retention and repelling negative pairs to prevent cross-concept interference. %
    }\vspace{-0.7em}
\label{fig:framework}
\end{figure*}

\section{Related Work}
\label{sec:related}

\noindent\textbf{Text-Conditioned Image Synthesis.}  
Text-conditioned image synthesis has seen significant advances through the development of both GANs~\cite{goodfellow2020generative, brock2018large, Karras2021, karras2019style, karras2020analyzing} and diffusion models~\cite{ddpm, song2020denoising, dhariwal2021diffusion, ho2022classifier, rombach2022high, nichol2021improved}. Early GAN-based methods focused on generating images conditioned on classes~\cite{brock2018large, karras2019style, huang2022multimodal} or text attributes~\cite{abdal2022clip2stylegan, gal2022stylegan, parmar2022spatially, roich2022pivotal}. More recently, the focus has shifted towards large-scale text-to-image diffusion models~\cite{rombach2022high, imagen, yu2022scaling, dalle2}, trained on large-scale datasets~\cite{schuhmann2022laion}, enabling more nuanced and accurate image synthesis.

\vspace{2pt}

\noindent\textbf{Personalized Image Generation and Customization.} 
Personalized image generation aims to embed user-specific concepts that can be reused across different contexts, from distinct characters to unique styles. Early methods like Textual Inversion (TI)~\cite{gal2022image} and DreamBooth (DB)~\cite{ruiz2023dreambooth} laid the groundwork by learning representations from a limited set of images. TI optimizes text embedding to reconstruct target images using a diffusion-based loss, allowing for flexible and personalized image synthesis. DB, on the other hand, fine-tunes model weights to learn unique concept representations, using rare tokens to encode custom features for reliable reproduction.
Subsequent works, such as $\mathcal{P}+$ \cite{voynov2023p+}, build upon Texture Inversion by incorporating a more expressive token representation, enhancing subject alignment and fidelity in generation.
Further developments have aimed to improve the scalability and efficiency of customization. Custom Diffusion~\cite{kumari2023multi} advanced this goal by fine-tuning only the cross-attention layers, balancing customization precision and computational efficiency. Building on these foundations, DB-LoRA~\cite{ryu2023low} introduces LoRA \cite{hu2021lora} to DB to enable more parameter-efficient tuning, reducing the need for extensive retraining. Recent approaches, such as StyleDrop~\cite{sohn2023styledrop}, HyperDreamBooth~\cite{ruiz2024hyperdreambooth}, and a variety of feed-forward network-based techniques~\cite{jia2023taming, ye2023ip-adapter, Wang2023StyleAdapterAS, Shi2023InstantBoothPT}, have further minimized computational demands by predicting adaptation parameters directly from data. 

\vspace{2pt}

\noindent\textbf{Merging Multiple Concepts.}  
Combining LoRAs for style and subject control remains an open research challenge. Current methods for multi-concept synthesis often face notable limitations. Weighted summation~\cite{ryu2023low} is a simple approach but suffers from feature interference. 
AnyDoor~\cite{chen2024anydoor} enables zero-shot object replacement and composition without per-instance fine-tuning.   Mix-of-Show~\cite{gu2023mix} requires specialized Embedding-Decomposed LoRAs (ED-LoRAs) for each concept, which limits compatibility with standard LoRAs. ZipLoRA~\cite{shah2023ziplora} can merge style and content LoRAs but struggles when multiple content LoRAs are needed. OMG~\cite{kong2024omg} depends on off-the-shelf segmentation to isolate subjects, making its performance highly dependent on segmentation accuracy and the model's ability to generate multiple objects. Recent advancements have focused on merging multiple specialized models into a unified generative framework. Mix-of-Show~\cite{gu2023mix} effectively merges models using ED-LoRAs but requires access to the original training data, which limits compatibility with community LoRAs, such as those available on platforms like civit.ai~\cite{civitai_website}. Orthogonal Adaptation~\cite{po2024orthogonal} introduces constraints to separate attributes across LoRAs, reducing interference; however, it increases training complexity by directly modifying the fine-tuning process, which also requires access to the original data.

Our approach, \model{}, uses a contrastive objective to align the weight spaces of specialized LoRA models, enabling coherent multi-concept compositions with minimal interference. Unlike prior methods, \model{} combines pre-existing LoRA models without retraining, preserving each model's distinct attributes for scalable, high-fidelity multi-concept image synthesis.

\section{Method}
\label{sec:method}

Our proposed approach, \model{}, enables seamless multi-concept synthesis by merging independently trained LoRA models in a post-training phase. Instead of modifying or retraining each model for compatibility, \model{} uses an optimization-based merging that adapts pre-existing LoRA models to function cohesively within a shared model, capable of generating all the target concepts. Leveraging contrastive learning, our method aligns the weight spaces of the models, ensuring each concept retains high-fidelity while remaining compatible in joint compositions. An overview of \model{} is shown in \cref{fig:framework}.

\subsection{Low-Rank Adaptation Models}

LoRA \cite{hu2021lora} fine-tunes large models by adding low-rank matrices \( W_{\text{in}} \) and \( W_{\text{out}} \) to frozen base layers, allowing adaptation with minimal computation. Applied to cross-attention layers in Stable Diffusion \cite{rombach2022high}, LoRA updates weights as \( W' = W + W_{\text{in}} W_{\text{out}} \), where \( W_{\text{in}} \) and \( W_{\text{out}} \) are significantly smaller than \( W \), reducing model storage to just 15-100 MB compared to the full model’s 3.44 GB. This fine-tuning embeds new styles or concepts efficiently within the diffusion model’s latent space.

\subsection{\textbf{\model{}}}

\model{}’s merging process relies on contrastive loss to ensure compatibility across independently trained LoRA models. Given a pre-trained LoRA $V_i$, we first create pairs of \textit{input} and \textit{output features}, denoted as \(X_i\) and \(Y_i\), respectively, see \cref{fig:framework} (a). %
Meanwhile, \textit{predicted features} \(\hat{Y_i}\) represent the merged model's output for the same input features, \(X_i\), \ie{,} \( \hat{Y_i} = (W + \Delta W) X_i \).
The intuition behind our contrastive objective is as follows:  positive pairs generated by the same LoRA models should attract, while negative pairs generated by the different LoRAs should repel, see \cref{fig:framework} (b). %
The contrastive loss objective is defined as:

\vspace{-4pt}
\begin{equation}
\mathcal{L}_{\text{contrastive}} = \frac{1}{N} \sum_{i=1}^N \left( d_{p,i}^2 + \max(0, m - d_{n,i})^2 \right),
\end{equation}

\noindent where \( d_{p,i} \) represents the positive distance for each pair, \( d_{n,i} \) represents the negative distance, \( m \) is the margin parameter, which defines the minimum allowable distance for negative pairs to enforce separation and prevent feature overlap, and \( N \) is the number of concepts being combined.
The positive and negative components are defined as:

\vspace{-4pt}
\begin{equation}
d_{p,i} = \|Y_{i} - \hat{Y}_{i}\|_2,
\end{equation}

\noindent where \( d_{p,i} \) is the distance between the output features \( Y_{i} \) of the original LoRA model and the predicted features \( \hat{Y}_{i} \) from the merged model for the same concept.

\vspace{-4pt}
\begin{equation}
d_{n,i} = \min_{j \neq i} \|Y_{i} - \hat{Y}_{j}\|_2,
\end{equation}

\noindent where \( d_{n,i} \) is the negative distance, computed as the minimum distance between the output features \( Y_{i} \) for a given concept and the predicted features \(  \hat{Y}_{j} \) of unrelated concepts, with \( j \neq i \).
This contrastive objective keeps unrelated concepts distinct while aligning each concept’s output features with the merged model’s predictions, with negative pair loss $d_n$ further preventing concept interference and ensuring distinct representations for similar concepts.

\vspace{4pt}

\paragraph{Delta-Based Merging.} \model{} uses an additive delta, \(\Delta W\), to merge LoRA models without altering the base weights directly. Initialized to zero, we learn \(\Delta W\) to adjust the pre-trained weights, preserving the integrity of each model while ensuring compatibility. An L2 regularization term is applied to \(\Delta W\) to limit its magnitude, ensuring sparsity and minimal adjustments.
The optimization objective combines the contrastive merging loss, \(\mathcal{L}_{\text{contrastive}}\) and \(\mathcal{L}_{\text{delta}}\):

\vspace{-4pt}
\begin{equation}
\mathcal{L}_{\text{delta}} = \lambda_{\text{delta}} \| \Delta W \|_2,
\end{equation}

\noindent where \(\lambda_{\text{delta}}\) controls the trade-off between effective merging of concepts and maintaining sparsity in \(\Delta W\). The total objective is:

\vspace{-4pt}
\begin{equation}
\mathcal{L}_{\text{total}} = \mathcal{L}_{\text{contrastive}} + \mathcal{L}_{\text{delta}}.
\end{equation}

At each step, \(\Delta W\) is updated to minimize \(\mathcal{L}_{\text{total}}\) using gradient descent. Positive and negative samples contribute to the contrastive pairs, reinforcing distinct boundaries for each concept while ensuring cohesive alignment among merged concepts. %
This iterative process converges to an optimized weight configuration. By restricting updates to \(\Delta W\) rather than altering base weights directly, \model{} achieves balanced adaptation, preserving the integrity of each concept’s features while enabling seamless multi-concept merging within a unified model.

\begin{figure*}
    \centering
    \includegraphics[width=\linewidth]{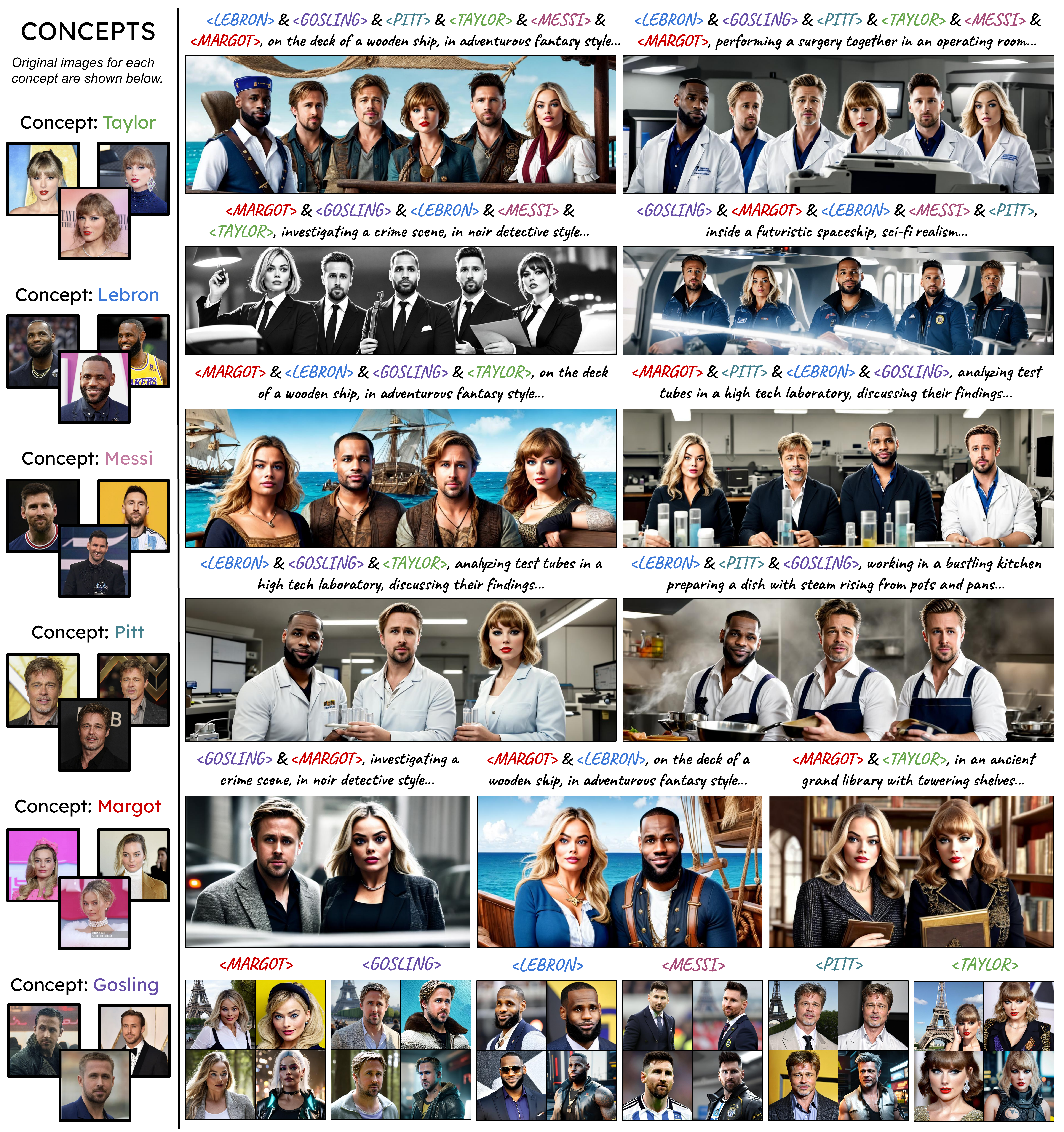}
\caption{\textbf{Qualitative Results.} \model{} effectively combines different numbers of unique concepts across a wide range of scenes, producing high-fidelity compositions that capture the complexity and nuance of multi-concept prompts in diverse environments. \model{} preserves the identity of each concept, ensuring accurate representation in composite scenes while also maintaining fidelity in single-concept generation, as demonstrated in the last row. Real images from the original concepts are shown on the left for reference.} \vspace{-10pt}
   \label{fig:qual}
\end{figure*}

\begin{figure*}
    \centering
    \includegraphics[width=\linewidth]{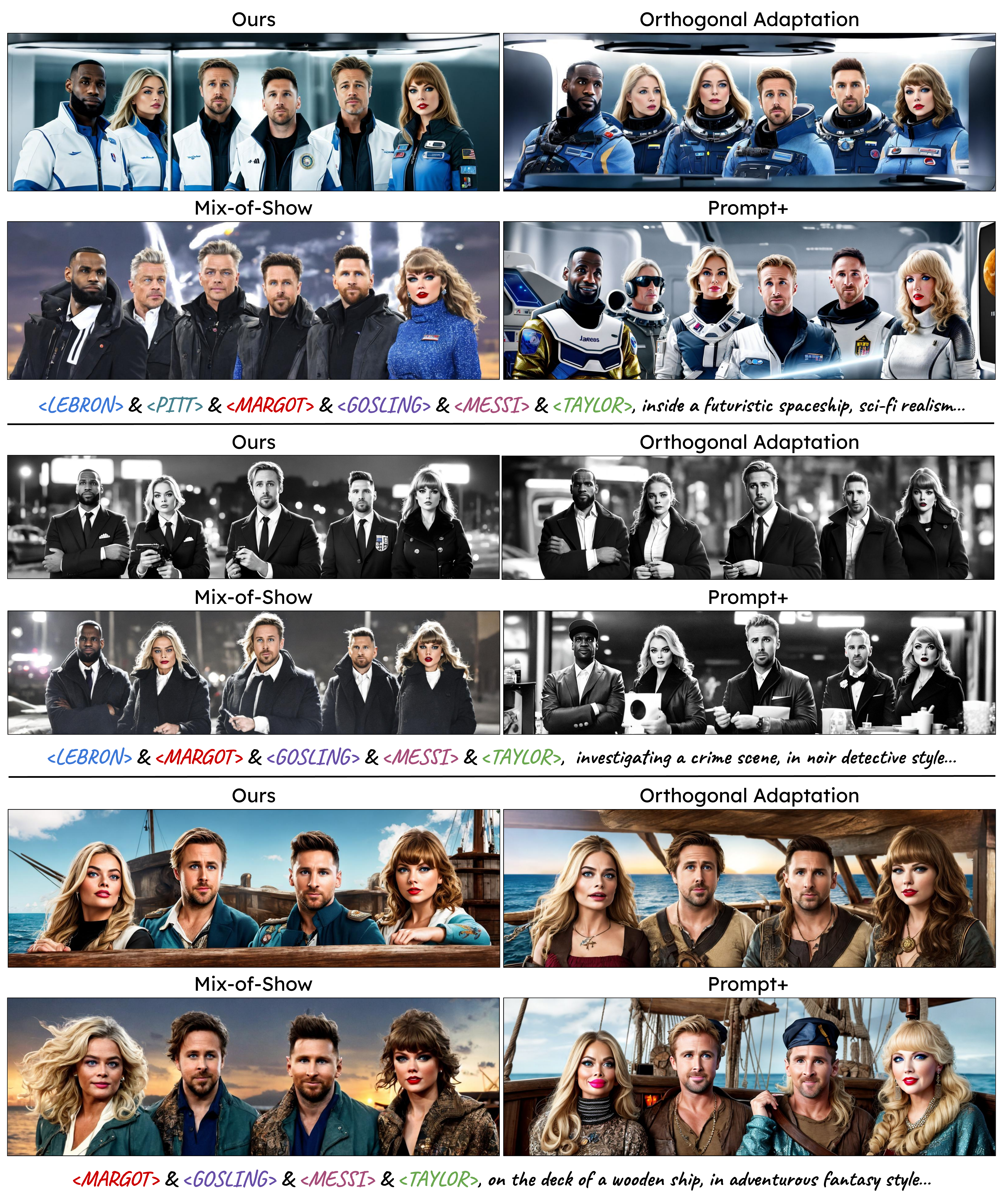}
    \vspace{-10pt}
    \caption{\textbf{Multi-Concept Comparison.} Composite images generated by our method (\model{}) and competing methods (Orthogonal Adaptation \cite{po2024orthogonal}, Mix-of-Show \cite{gu2023mix}, Prompt+ \cite{voynov2023p+}) for multi-concept prompts. Each row depicts a different scene defined by the text prompts. Our method consistently preserves individual identities, while others struggle with identity preservation and concept interference.}
    \label{fig:comparison} \vspace{-10pt}
\end{figure*}

\section{Experiments}
\label{sec:exp}

\paragraph{Implementation Details.} In all our experiments, we use the Stable Diffusion model~\cite{rombach2022high} with the ChilloutMix checkpoint\footnote{\url{https://huggingface.co/windwhinny/chilloutmix}} for its high-quality image generation capability. Our approach leverages pre-trained LoRA models, eliminating the need to train individual LoRA models from scratch and significantly reducing computational overhead. During the merging process, we use a learning rate of $1e^{-4}$, a margin parameter \(m\) of 0.5 to control separation between positive and negative pairs during contrastive learning, and a regularization coefficient $\lambda_{\text{delta}}$ of 0.001 to ensure sparsity in the learned delta. 
All experiments, including model outputs, and processing, 
were conducted on an NVIDIA A100 GPU at Virginia Tech.
Combining 12 concepts is highly efficient, typically taking about 5 minutes, which makes our approach scalable and practical for real-world applications that require seamless integration of multiple concepts. For our experiments, we utilize pre-trained single-concept LoRA and/or ED-LoRA models as the starting point. %

\paragraph{Baselines.} We evaluate our approach against several state-of-the-art baselines in multi subject customization, including: DB-LoRA~\cite{ruiz2023dreambooth}, $\mathcal{P}+$~\cite{voynov2023p+}, Custom Diffusion~\cite{kumari2023multi}, Mix-of-Show~\cite{gu2023mix}, and Orthogonal Adaptation (OA)~\cite{po2024orthogonal}.
For DB-LoRA, Federated Averaging (FedAvg) is used to merge models. Custom Diffusion utilizes its optimization-based method for merging, while Mix-of-Show merges using gradient fusion. Orthogonal Adaptation introduces orthogonal transformations to mitigate concept interference during merging. In the case of $\mathcal{P}+$, no fine-tuning of model weights is performed; instead, merging is achieved by directly querying each concept's token embedding.

\paragraph{Datasets \& Metrics.} 
For all evaluations, we use the same experimental setup and dataset proposed by \cite{po2024orthogonal}, which includes 12 concept identities, each represented by 16 distinct images of the target concept across various contexts.
Following prior work~\cite{po2024orthogonal, gu2023mix}, we evaluate our method using three key metrics: Text Alignment, Image Alignment, and Identity Alignment. Text Alignment measures text-image similarity using the CLIP \cite{radford2021learning} model to ensure that generated images match the input prompts. Image Alignment evaluates the similarity between generated and reference images in the CLIP feature space. Identity Alignment uses the ArcFace \cite{deng2019arcface} model to assess how accurately the target human identity is preserved in the generated images.

\subsection{Qualitative Results}

We demonstrate the qualitative results of our method for both single and multiple subjects. Given a set of subjects, each represented by a personalized model—for instance, celebrities like Margot Robbie and Taylor Swift (see \cref{fig:qual})—and a text prompt such as `\texttt{<Margot>} and \texttt{<Taylor>} in an ancient library with towering shelves,' our objective is to generate a composite image that integrates these subjects according to the provided text prompt. Using 12 subjects identified by \cite{po2024orthogonal}, we first merge them into a unified model through our novel contrastive-based objective, applied consistently across all experiments. Unlike methods that require individual fine-tuning for each concept \cite{po2024orthogonal}, our approach can work with pre-trained models such as LoRA or ED-LoRA models.

\paragraph{Single Concepts.} We first demonstrate our method's ability to preserve individual identities. \Cref{fig:qual} (bottom row) confirms that our approach maintains the integrity of each identity for single concepts.   This capability extends to diverse settings. For example, the ``\texttt{<CONCEPT>} Cyberpunk style' shows subjects portrayed as game-like characters, as seen in the fourth images of the bottom row in \cref{fig:qual}. %

\begin{figure*}[!htb]
\centering
\includegraphics[width=1\linewidth]{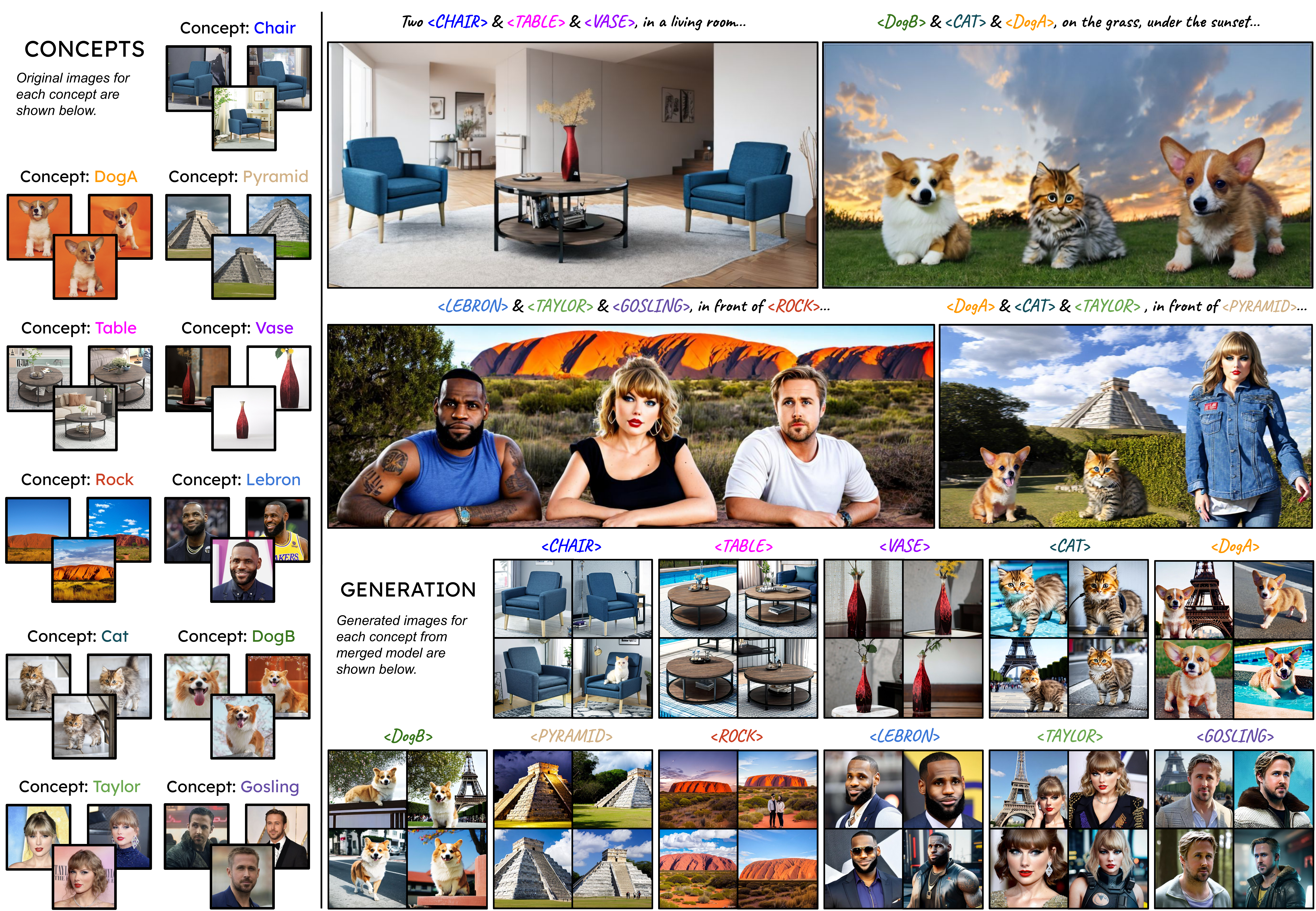} 
\caption{\textbf{Non-human subject \& Complex Composition.} (a) Our method effectively integrates diverse concepts—including animals, objects (\eg, tables, chairs, vases), and landmarks (\eg, pyramids, rocks)—into visually coherent and semantically meaningful scenes. (b) \model{} also enables complex interactions between concepts, such as multiple subjects engaging in contextual activities.}
\label{fig:nonhuman} 
\end{figure*}

\paragraph{Multiple Concepts.} \Cref{fig:qual} showcases images generated with our method using varying numbers of concepts—6, 5, 4, 3, and 2. These visuals demonstrate that our method not only accurately captures each individual identity but also generates composite images guided by the text prompts. Notably, our approach excels by utilizing a single merged model to generate a wide range of concepts across diverse prompts, eliminating the need to train separate models for different concept counts. %
We also note that the original LoRA models for individual subjects are trained on various images of celebrities who may have different hairstyles or colors. As a result, variations such as different hairstyles or colors can appear even within generations of the same celebrity model (\eg{,} different generations of a Taylor Swift LoRA might show varying hair colors or styles, reflecting her real-life changes). These variations are not inconsistencies in our model but rather reflect the diversity inherent in the original LoRA models. Importantly, despite these variations, our model maintains fidelity and identity to the celebrities' faces and ensures that details like hairstyles do not mix across different subjects.

\paragraph{Qualitative Comparisons.} In \cref{fig:comparison}, we compare our method against several state-of-the-art approaches, including OA, MoS, and $\mathcal{P}+$, across various concepts. \Cref{fig:comparison} illustrate that our method successfully preserves individual identities without crossovers, whereas other methods encounter issues. For instance, OA \cite{po2024orthogonal} inadvertently transfers features, such as the hair of female characters, to others, while MoS \cite{gu2023mix} and $\mathcal{P}+$ \cite{voynov2023p+} struggle with precise identity depiction. Additionally, while these models perform adequately with two or three concepts, their ability to accurately represent more concepts diminishes as the concept count increases. For instance, in scenes with six concepts, see \cref{fig:comparison} first section, methods other than \model{} fail to preserve the identity of Messi, highlighting their limitations as the number of concepts increases.  %

\begin{figure}[!htb]
\centering
\includegraphics[width=\linewidth]{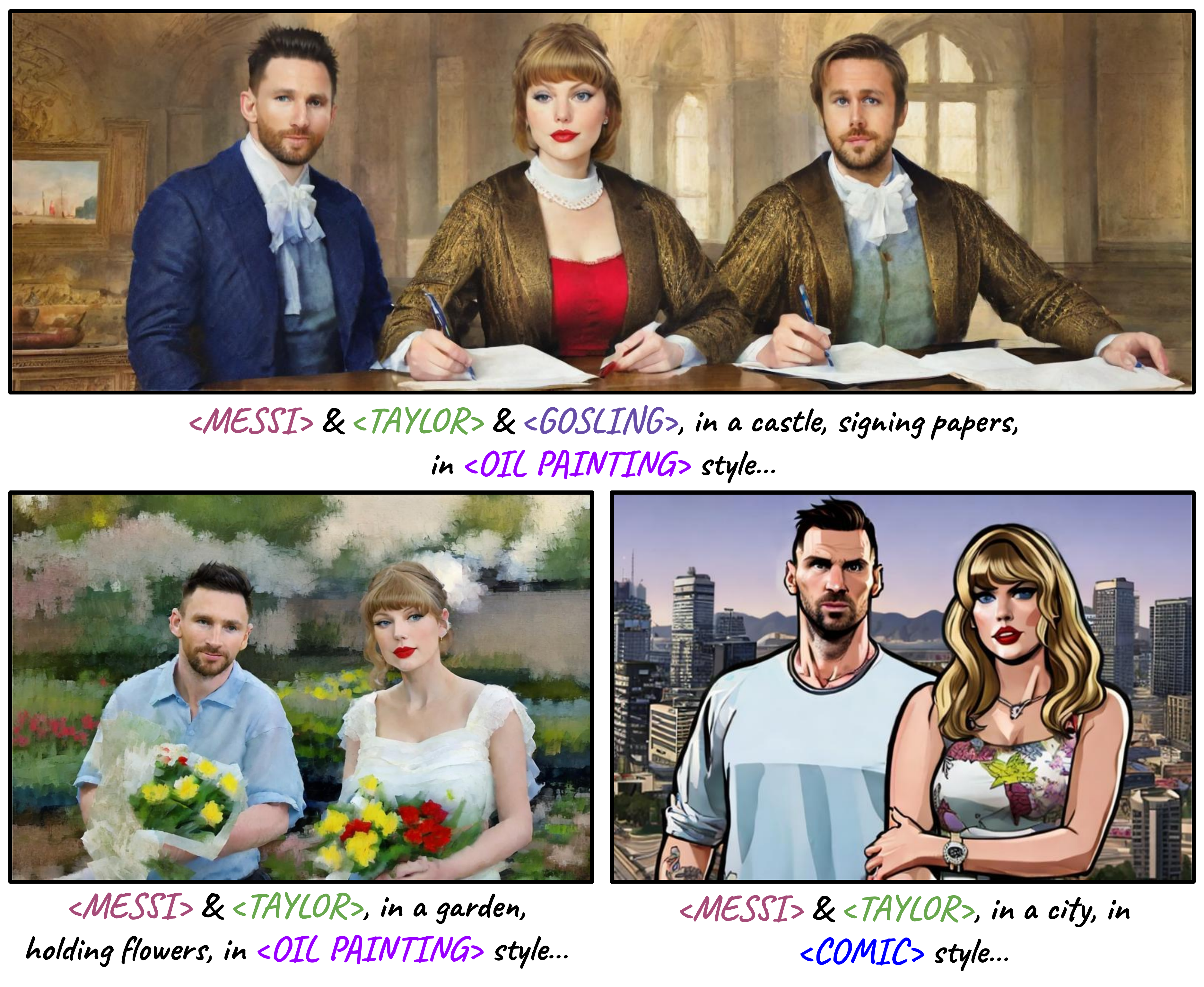}
\caption{\textbf{Style LoRAs.} Our method combines styles like comic art and oil painting into multi-subject scenes, ensuring stylistic fidelity and content coherence and showcasing its flexibility.}
\label{fig:style} 
\end{figure}

\paragraph{Style LoRA Integration.}
To showcase the flexibility of our approach, we integrate style-specific LoRA models to generate scenes combining conceptual and stylistic elements, enabling outputs in styles like \textit{comic art} or \textit{oil painting}. As shown in \cref{fig:style}, prompts such as ``...in a city, in \textit{comic} style" capture the vibrant aesthetic of comic art, while ``...in a garden, holding flowers, in {oil painting} style" reflect the textured quality of oil painting. Complex scenes like ``...in a castle, signing papers, in {oil painting} style" illustrate the model's ability to maintain content coherence alongside stylistic adaptation. These results demonstrate our method's efficiency in preserving content accuracy and achieving high stylistic fidelity, making it practical for creative and artistic workflows.

\begin{table*}[hbt!]
  \centering
    \caption{\textbf{Quantitative Results.} Comparison of \model{} against state-of-the-art models, evaluated before and after merging. \model{} achieves competitive performance across all metrics, with notable improvements in image and identity alignment post-merging.}
  \label{tab:quantitative_eval}
    \resizebox{0.98\textwidth}{!}{

  \begin{tabular}{lccccccccc}
  \toprule
    \multirow{2}{*}[-0.1cm]{Method} & \multicolumn{3}{c}{Text Alignment $\uparrow$} & \multicolumn{3}{c}{Image Alignment $\uparrow$} & \multicolumn{3}{c}{Identity Alignment $\uparrow$} \\
    \cmidrule{2-10}
    & Single  & Merged  & $\Delta$ & Single  & Merged  & $\Delta$ & Single  & Merged  & $\Delta$\\
    \midrule
    P+~\cite{voynov2023p+} & .643\RA & .643 & --- & .683\RA & .683 & --- & .515\RA & .515 & --- \\
    Custom Diffusion~\cite{kumari2023multi} & .668\RA & .673 & +.005 & .648\RA & .623 & -.025 & .504\RA & .408 & -.096 \\
    DB-LoRA (FedAvg)~\cite{ryu2023low} & .613\RA & .682& +.069 & .744\RA & .531 & -.213 & .683\RA  & .098 & -.585 \\
    MoS (FedAvg)~\cite{gu2023mix} & .625\RA & .621 & -.004 & .745\RA & .735 & -.010 & .728\RA & .706 & -.022 \\
    MoS (Grad Fusion)~\cite{gu2023mix} & .625\RA & .631 & +.006 & .745\RA& .729 & -.016 & .728\RA & .717 & -.011 \\
    Orthogonal Adaptation~\cite{po2024orthogonal} & .624\RA & .644 & +.020 & .748\RA & .741 & -.007 & .740\RA & .745 & +.005\\ \midrule
    \model{} (Ours) & .668\RA & .665 & -.003 & .766\RA & \textbf{.776} & +.010 & .799\RA & \textbf{.828} & +.029\\
    \bottomrule
  \end{tabular}}
\label{tab:comp}
\end{table*}

\begin{figure*}[!htb]
    \centering
    \includegraphics[width=1\linewidth]{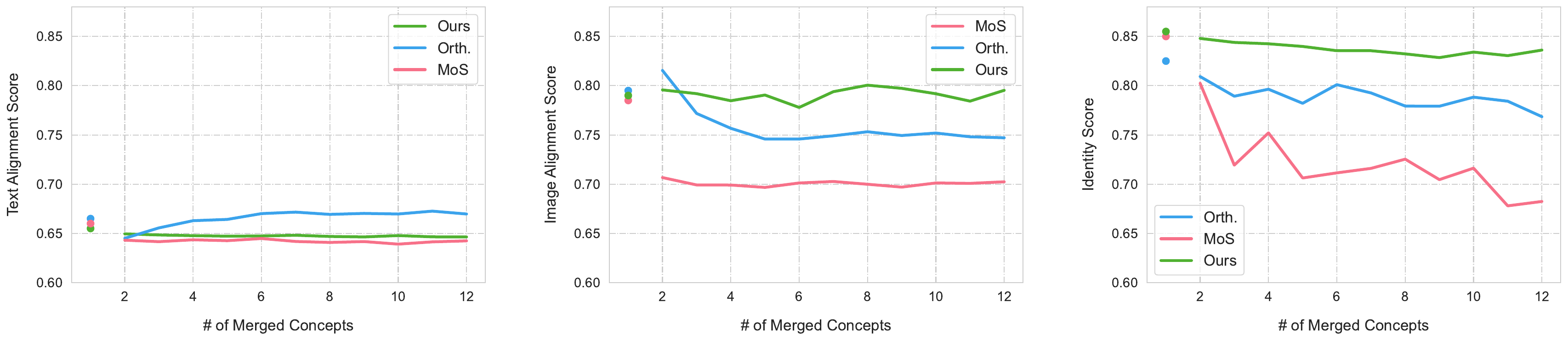}

        \caption{\textbf{Quantitative Results on Number of Concepts.} Text alignment, image alignment, and identity preservation scores remain high as merged concepts increase. Our method maintains identity and prompt adherence. Dots indicate baseline metrics for each LoRA model pre-merging for performance reference.}
    \label{fig:ablation_n}
\end{figure*}

\paragraph{Non-Human Examples.}
We present examples involving non-human concepts, including animals, objects, and monuments. Our model generalizes seamlessly to these scenarios while maintaining high-quality outputs and stylistic consistency.
As shown in \cref{fig:nonhuman}, our method effectively integrates diverse concepts into cohesive scenes. It excels with objects like tables, chairs, and vases, accurately capturing their distinct textures. Additionally, it demonstrates robustness with monumental elements such as pyramids and rocks, blending them seamlessly with other concepts.
These results highlight the adaptability of our method in generating coherent and visually engaging compositions across a wide range of non-human concepts, including animals, objects, and landmarks. This versatility makes it suitable for both creative and practical applications, such as wildlife illustration, interior design, and architectural visualization.

\subsection{Quantitative Results}
In \cref{tab:comp}, we present the results for each approach before and after identity merging to illustrate their impact on the metrics. Our approach achieves the highest scores for image and identity alignment compared to other methods, as demonstrated by the qualitative examples in \cref{fig:comparison}. While Custom Diffusion and DB-LoRA show superior Text Alignment, they fall short in Image and Identity Alignment, underscoring the versatility and balanced performance of our approach across all key aspects.

\paragraph{Effect of Number of Concepts.} 
Unlike other methods, which struggle to maintain text alignment, image alignment, and identity preservation as the number of combined concepts increases, \model{} preserves these metrics. As shown in Fig. \ref{fig:ablation_n}, our method remains effective even as the number of concepts grows.

\paragraph{User Study.} To supplement the findings in \cref{tab:comp} and \cref{fig:ablation_n}, we conduct a user study involving 50 participants on Prolific.com  \cite{prolific}. The study consisted of 40 questions where participants are shown reference images of individual concepts alongside composite images,  with varying numbers of concepts, generated by our method and competing methods, presented in a randomized order. Participants are asked to evaluate each image pair based on: \textit{Identity Alignment: Given the reference images on the left, how well does the image on the right capture the identity of these concepts?} (1 = Not at all, 5 = Very much). %
Our method achieves significantly higher ratings for identity alignment compared to other methods, indicating its superior ability to maintain concept identity in composite images, see \cref{tab:user}. %

\begin{table}[!htb]
    \centering
    \caption{\textbf{User Study Results.} Our method achieves the highest average score for identity alignment, indicating superior preservation of concept identities compared to competing methods.}
    \label{tab:user}
    \resizebox{0.85\linewidth}{!}{
        \begin{tabular}{cc}
            \toprule
             Method & Identity Alignment \\ \midrule
             Ours & \textbf{3.42} \\
             Orthogonal Adaptation & 2.41 \\
             Mix-of-Show & 2.21 \\
             Prompt+ & 2.01 \\ \bottomrule
        \end{tabular}
    }
\end{table}

\paragraph{Merging Time.} In terms of time efficiency, our method demonstrates significant advantages. It takes only 5 minutes to combine 12 LoRA models. In contrast, \cite{po2024orthogonal} requires fine-tuning each LoRA model from scratch, each taking approximately 10-15 minutes. While the actual merging process of \cite{po2024orthogonal} takes only 1 second, it requires prior fine-tuning, which adds up to a total of 120 minutes to merge them.
Meanwhile, Mix-of-Show requires 15 minutes to merge the models. After merging, generating the images takes approximately  10 seconds for all methods. This comparison clearly shows that our method is substantially faster than other methods in merging models. Once \model{} merges LoRA models into a unified model, it can generate composite images without any further need for retraining the individual concepts or access to original training data.

\begin{figure}[!ht]
    \centering
    \includegraphics[width=\linewidth]{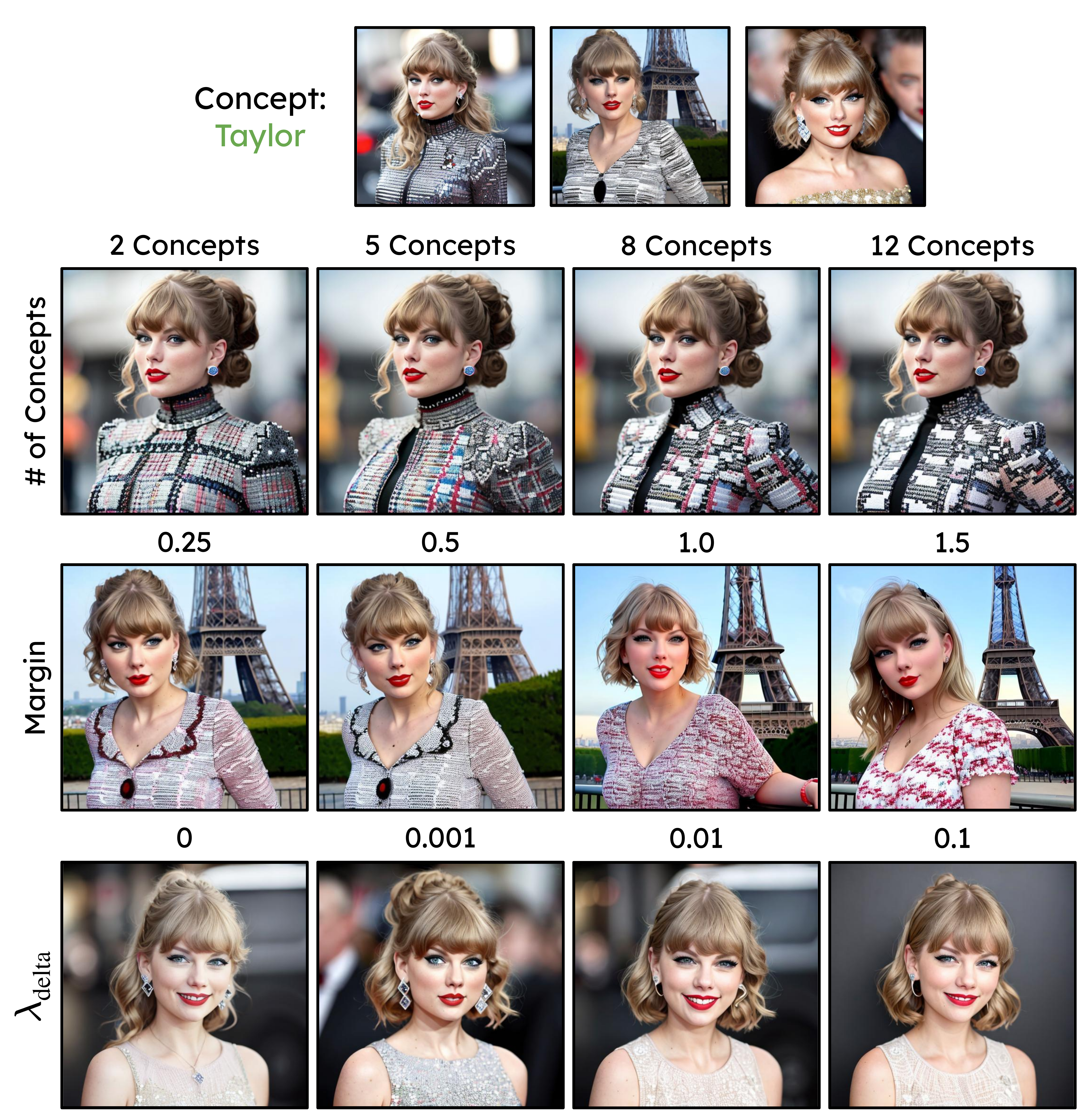}
    \caption{\textbf{Ablation Study on Margin, $\lambda_{\text{delta}}$, and Concept Count.} Effect of varying margin, $\lambda_{\text{delta}}$, and number of concepts (2, 5, 8, 12) on identity preservation and visual coherence. }
    \label{fig:ablation}
\end{figure}

\paragraph{Ablation Study.}
We conduct ablation studies to evaluate the impact of key parameters such as the effects of margin, $\lambda_{\text{delta}}$, and concept count as follows:
\paragraph{(i) Impact of Margin and $\lambda_{\text{delta}}$.} As shown in \cref{fig:ablation}, we explore different values of margin and $\lambda_{\text{delta}}$. The results demonstrate that our method achieves robust identity preservation and visual coherence with margin values around 0.25–0.5 and $\lambda_{\text{delta}}$ set at 0.001. Higher values for either parameter lead to diminished performance in maintaining individual identity and prompt coherence.
\paragraph{(ii) Effect of Number of Concepts.} Qualitative results in \cref{fig:ablation} further demonstrate that our model maintains identity and visual coherence even with complex multi-concept compositions, highlighting its scalability and robustness.

\section{Discussion}

\paragraph{Limitations.} Similar to  \cite{gu2023mix, po2024orthogonal}, the performance of our method is inherently tied to the capabilities of the underlying LoRA models. Therefore, the success of \model{} in generating coherent and high-quality images depends on the robustness  of these initial models. Moreover, given that our method enables sophisticated composition capabilities, it is crucial to consider the potential for misuse, such as  deepfakes. Thus, we advocate for the careful use of our method to prevent such applications and  promote ethical use.

\paragraph{Conclusion.} We introduce \model{}, a novel approach that merges multiple LoRA models using a contrastive learning objective. Our method preserves the distinct identities of concepts while enabling the creation of composite images from multiple subjects. \model{} is designed to be compatible with any existing LoRA model, ensuring seamless integration with future, more advanced models.  Operating as a post-training method, \model{} requires only a one-time merging process and can be used with any prompts, utilizing community-available LoRA models. %

{
    \small
    \bibliographystyle{ieeenat_fullname}
    \bibliography{main}
}

\clearpage
\maketitlesupplementary

\section{Evaluation for Multi-concept Generation}
Although we follow the same evaluation setup as other competitors, we acknowledge that instance-based metrics may better capture accuracy and identity for multi-concept image generation. To further supplement our current metrics, we develop a pipeline to quantify composition accuracy and identity. 
Our pipeline begins by segmenting individual subjects using the model from Li et al. \cite{li2024omgseg}, allowing us to isolate key elements in both generated and ground truth images. Once segmented, each subject is assigned to a corresponding concept based on extracted feature similarities, ensuring a meaningful comparison between generated outputs and reference subjects. We then compute three instance-based metrics to evaluate the composition. Accuracy measures correct subject reconstruction by comparing segmentation masks, while Identity quantifies how well the generated subject preserves its identity using DINO-based feature similarity, which captures fine-grained semantic consistency. Additionally, CLIP-I Similarity assesses alignment between generated subjects and their ground truth counterparts using CLIP embeddings.
\model{} demonstrates significant improvements across these metrics, reinforcing our qualitative findings, see \cref{tab:instance-quantitative}.

\begin{table}[!htb]
    \centering
    \caption{\textbf{Quantitative comparison of instance-based metrics.} Our approach achieves higher scores across all metrics, indicating better subject disentanglement and identity preservation. \label{tab:instance-quantitative}} 
\resizebox{0.9\linewidth}{!}{%
\begin{tabular}{lcccc}
\toprule
                                          & Method      & Average          & Min           & Max           \\ \midrule
                     \multirow{4}{*}{\rotatebox[origin=c]{90}{%
            \begin{tabular}{@{}c@{}}
                Accuracy
            \end{tabular}%
        }} & \textbf{Ours}        & \textbf{0.724 ± 0.042} & \textbf{0.265 ± 0.044} & \textbf{0.918 ± 0.027} \\
                      & Orthogonal   & 0.684 ± 0.048 & 0.200 ± 0.040 & 0.832 ± 0.042 \\
                      & Mix-of-Show & 0.659 ± 0.036 & 0.163 ± 0.037 & 0.855 ± 0.037 \\
                      & Prompt +    & 0.582 ± 0.039 & 0.102 ± 0.030 & 0.816 ± 0.039 \\ \midrule                     \multirow{4}{*}{\rotatebox[origin=c]{90}{%
            \begin{tabular}{@{}c@{}}
                Identity
            \end{tabular}%
        }} & \textbf{Ours}        & \textbf{0.731 ± 0.891} & \textbf{0.617 ± 0.083} & \textbf{0.849 ± 0.105} \\
                      & Orthogonal   & 0.708 ± 0.142 & 0.593 ± 0.152 & 0.828 ± 0.148 \\
                      & Mix-of-Show & 0.666 ± 0.143 & 0.543 ± 0.158 & 0.804 ± 0.136 \\
                      & Prompt +    & 0.631 ± 0.122 & 0.519 ± 0.116 & 0.783 ± 0.132 \\ \midrule
                     \multirow{4}{*}{\rotatebox[origin=c]{90}{%
            \begin{tabular}{@{}c@{}}
                CLIP-I
            \end{tabular}%
        }} & \textbf{Ours}        & \textbf{0.722 ± 0.061} & \textbf{0.495 ± 0.074} & \textbf{0.860 ± 0.054} \\
                      & Orthogonal   & 0.698 ± 0.066 & 0.475 ± 0.073 & 0.834 ± 0.059 \\
                      & Mix-of-Show & 0.652 ± 0.075 & 0.459 ± 0.069 & 0.803 ± 0.076 \\
                      & Prompt +    & 0.650 ± 0.071 & 0.455 ± 0.072 & 0.801 ± 0.061 \\ \midrule
            \multirow{4}{*}{\rotatebox[origin=c]{90}{%
                \begin{tabular}{@{}c@{}}
                    DINO
                \end{tabular}%
            }}    & \textbf{Ours}        & \textbf{0.510 ± 0.052} & \textbf{0.189 ± 0.088} & \textbf{0.771 ± 0.041} \\
                                          & Orthogonal   & 0.502 ± 0.101 & 0.181 ± 0.092 & 0.749 ± 0.043 \\
                                          & Mix-of-Show & 0.412 ± 0.042 & 0.141 ± 0.051 & 0.681 ± 0.081 \\
                                          & Prompt +    & 0.445 ± 0.061 & 0.137 ± 0.066 & 0.703 ± 0.105 \\ 
                                          \bottomrule
\end{tabular}
}
\end{table}

\section{Our Approach with SOTA}
\begin{figure}[!htb]
    \centering
    \includegraphics[width=0.9\linewidth]{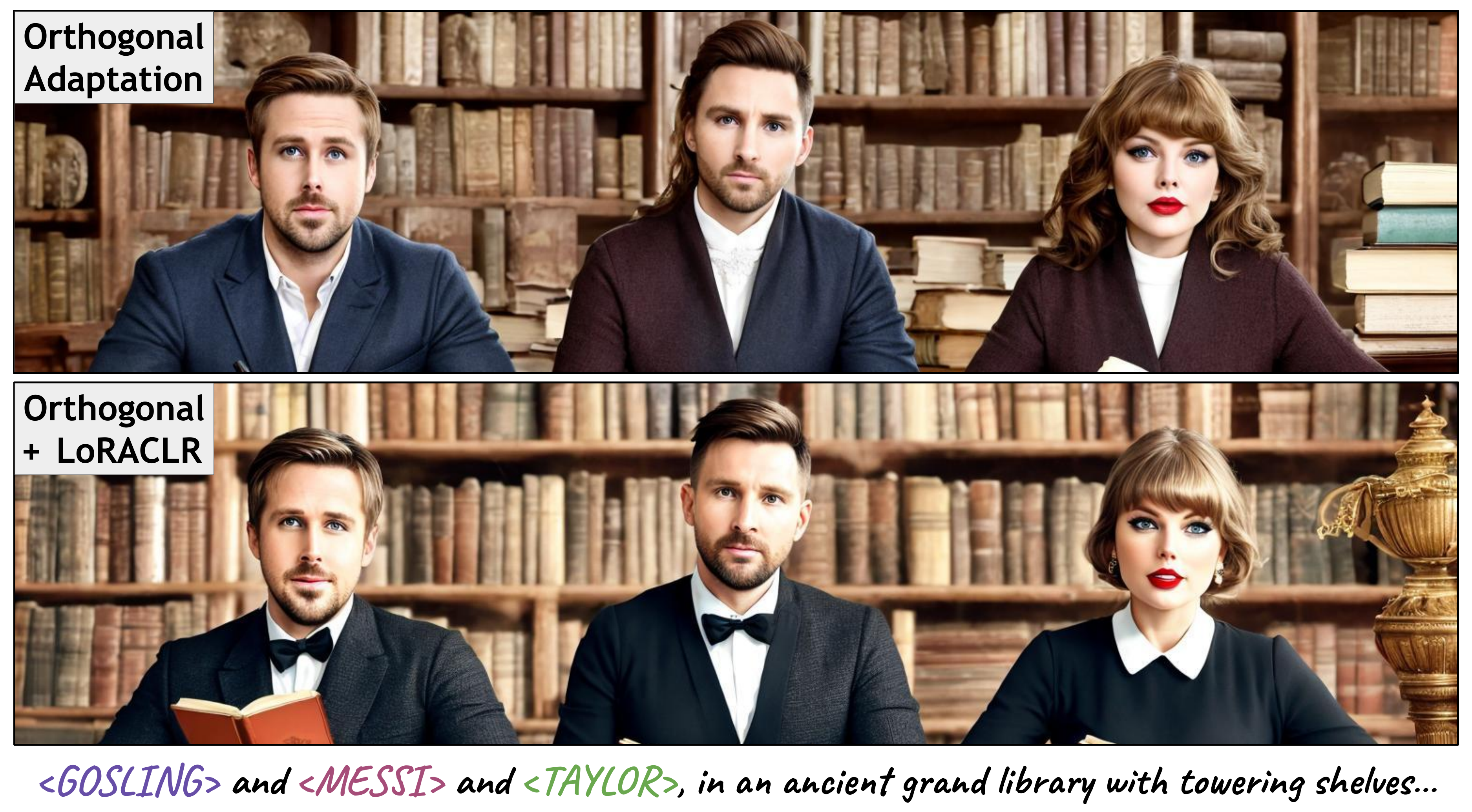}
    \vspace{-10pt}
    \caption{\textbf{Combining \model{} with SOTA.} Our method can be combined with other methods to reduce interference, resolving cases like Messi’s hair blending with Taylor’s, demonstrating its ability to enhance existing SOTA methods.} 
    \label{fig:ortholoraclr}
\end{figure}

We extend our implementation to incorporate state-of-the-art (SOTA) methods, such as Orthogonal Adaptation. By integrating our approach, we further enhance the disentanglement capabilities of these methods while mitigating concept interference. As illustrated in \cref{fig:ortholoraclr}, \model{} effectively resolves issues such as Messi’s hair blending with Taylor’s, demonstrating its ability to refine concept separation in challenging cases. These findings highlight \model{}’s potential to further improve existing SOTA, making it a valuable complement to current techniques.

\section{Concept Interactions}  

\begin{figure}[!htb]
    \centering
    \includegraphics[width=0.9\linewidth]{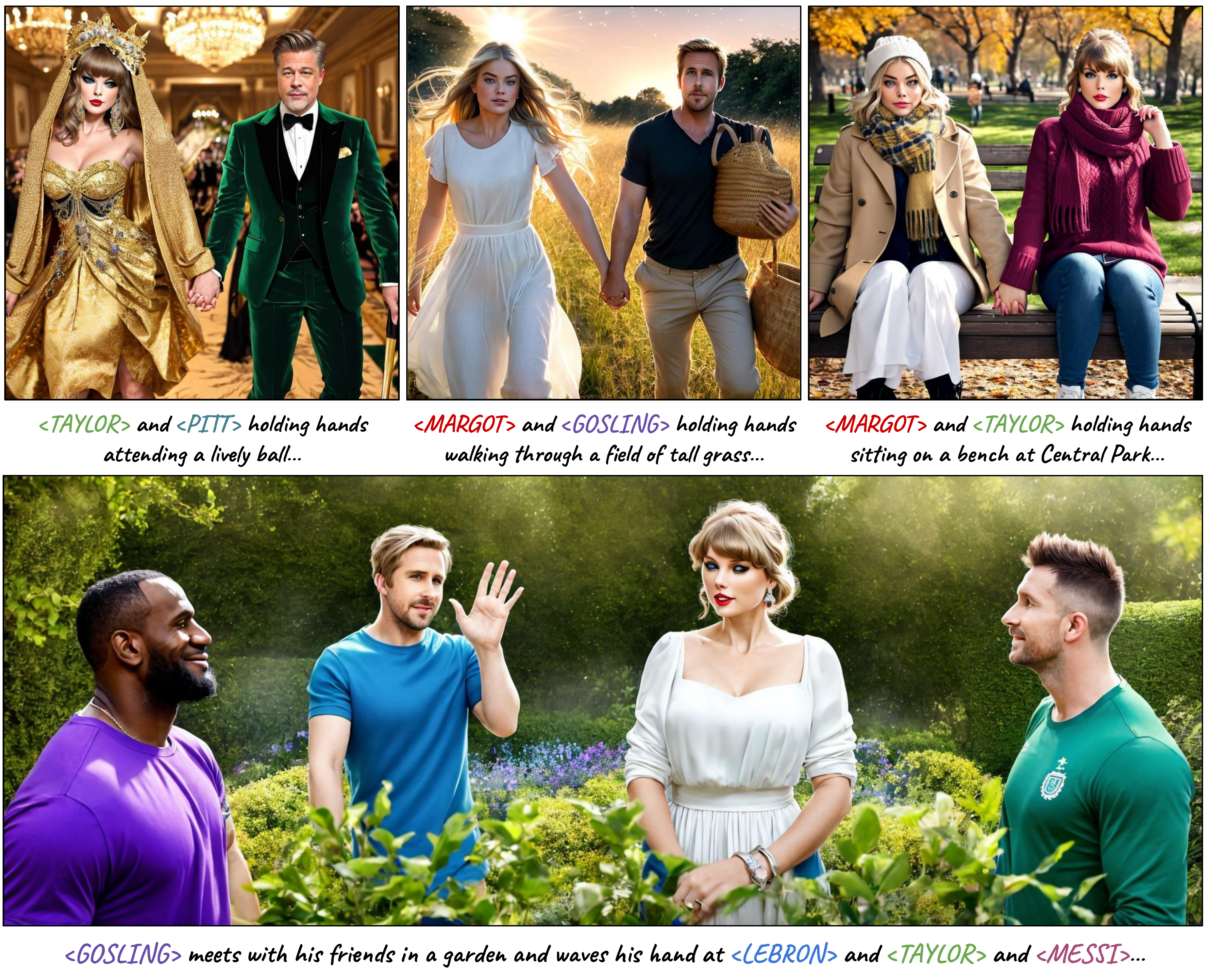}
    \vspace{-10pt}
    \caption{\textbf{Interactions between Concepts.} Examples of subjects holding hands and waving, showcasing our method's ability to generate coherent multi-concept compositions.}    \label{fig:concept-interactions}
\end{figure}

Our method is capable of handling interactions between multiple concepts, such as "holding hands" and "waving," ensuring coherent composition and spatial relationships. As shown in \cref{fig:concept-interactions}, our approach successfully generates complex interactions between subjects while maintaining realism and consistency.

\begin{figure*}[!htb]
    \centering
    \includegraphics[width=0.98\linewidth]{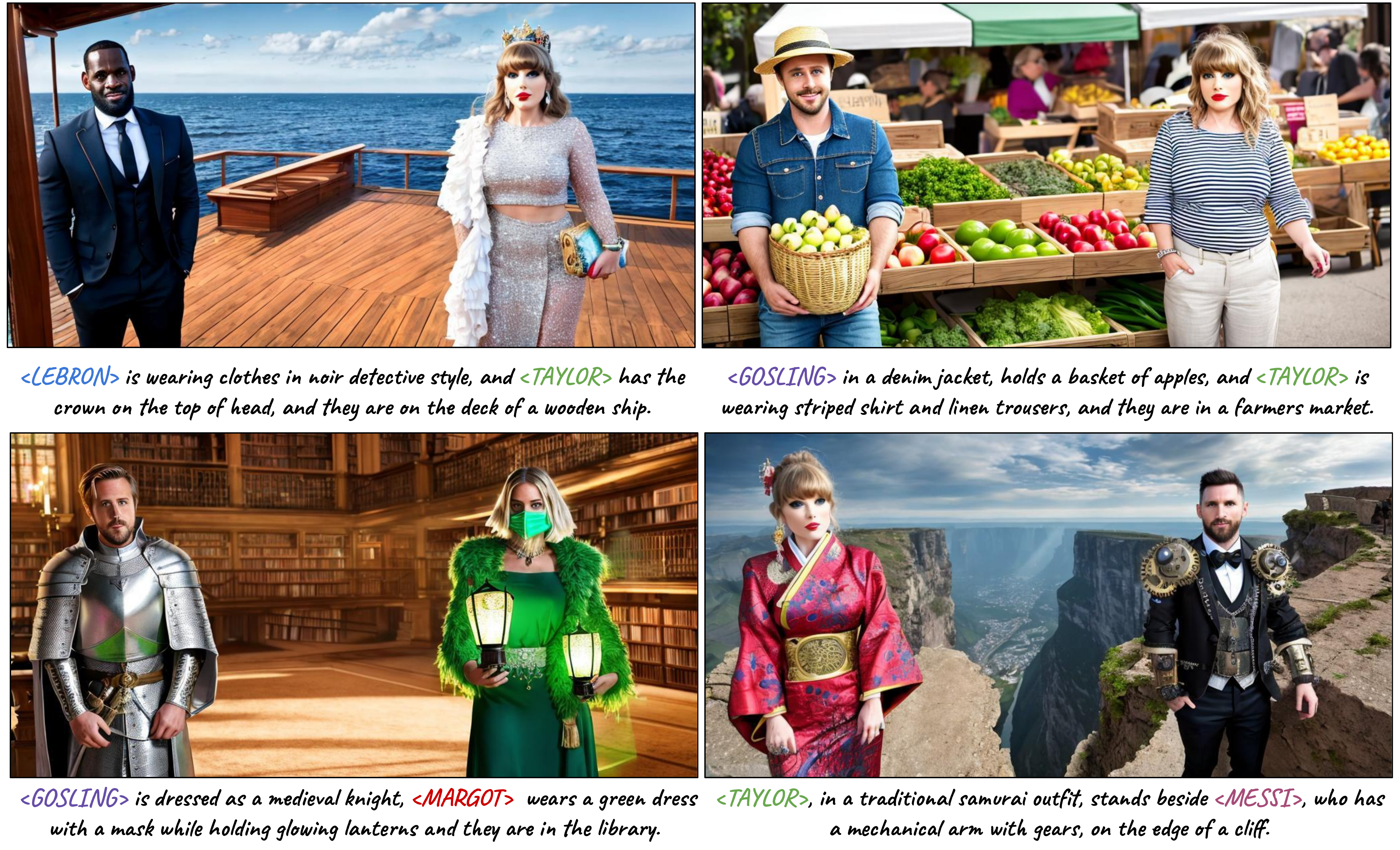}
    \vspace{-10pt}
    \caption{\textbf{Diverse Prompt Configurations.} Examples demonstrating the flexibility of our approach in multi-concept generation. Concepts are modified with different attributes, such as styles, objects, and accessories, enabling more controlled and varied outputs.}    \label{fig:flex-prompt}
\end{figure*}

\begin{figure}[!htb]
    \centering
    \includegraphics[width=\linewidth]{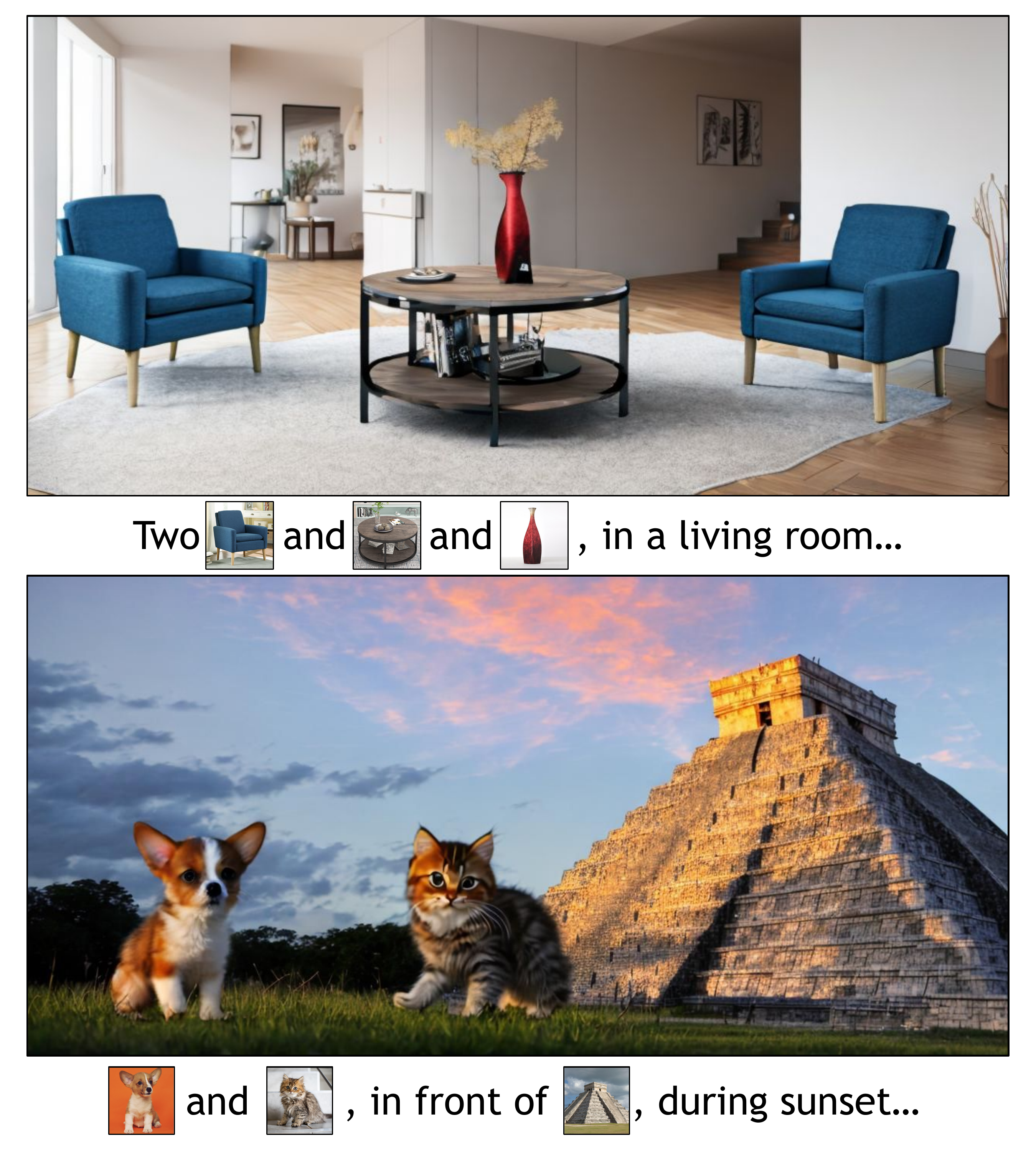}
    \vspace{-10pt}
    \caption{\textbf{Diverse Concept Placement.} Examples of stacked and back-to-back object configurations, demonstrating our method’s ability to generate flexible spatial arrangements.}
    \label{fig:placement-diversity}
\end{figure}

\section{Concept Placement Diversity}  

Our method effectively handles diverse and non-linear concept configurations, enabling flexible spatial arrangements. As shown in \cref{fig:placement-diversity}, our approach successfully generates compositions where objects are stacked, such as a vase placed on top of a chair, or positioned back-to-back, such as a dog and a cat in front of a pyramid. This demonstrates our model's ability to preserve spatial relationships while maintaining visual coherence.

\section{Flexibility in Prompt Setup}
We use the same setup as Orthogonal Adaptation for prompts (\eg, $\langle$V1$\rangle$, $\langle$V2$\rangle$, $\langle$V3$\rangle$...), but {our method is flexible and accommodates different prompt configurations}. As shown in \cref{fig:flex-prompt}, our approach allows generating concepts with various objects or accessories, such as "\textit{$\langle$V1$\rangle$ in noir style}" or "\textit{$\langle$V2$\rangle$ with a crown}". This flexibility enables more diverse and controllable multi-concept compositions.

\section{More Comparison}
In addition to the comparisons presented in the main paper, this section highlights further evaluations to emphasize the robustness of our method.

\subsection{Comparison with OMG}
OMG~\cite{kong2024omg} relies on a two-step process for scene generation. First, it generates a layout that structures the composition of the scene. Next, it populates this layout by placing the subjects in their respective positions. This dependency on intermediate layout generation introduces notable limitations. Errors in the layout creation stage often propagate, resulting in inconsistencies in the final output. Additionally, OMG struggles with scenarios involving subjects that share similar attributes, such as two individuals of the same gender (\eg{,} two women). This limitation leads to reduced quality and coherence in the generated images. Furthermore, since OMG operates in two stages, it requires approximately twice the inference time compared to single-stage approaches, \eg{,} ours, Mix-of-show and Orthogonal Adaptation, making it less efficient for real-time or large-scale applications.

\begin{figure*}
    \centering
    \includegraphics[width=1\linewidth]{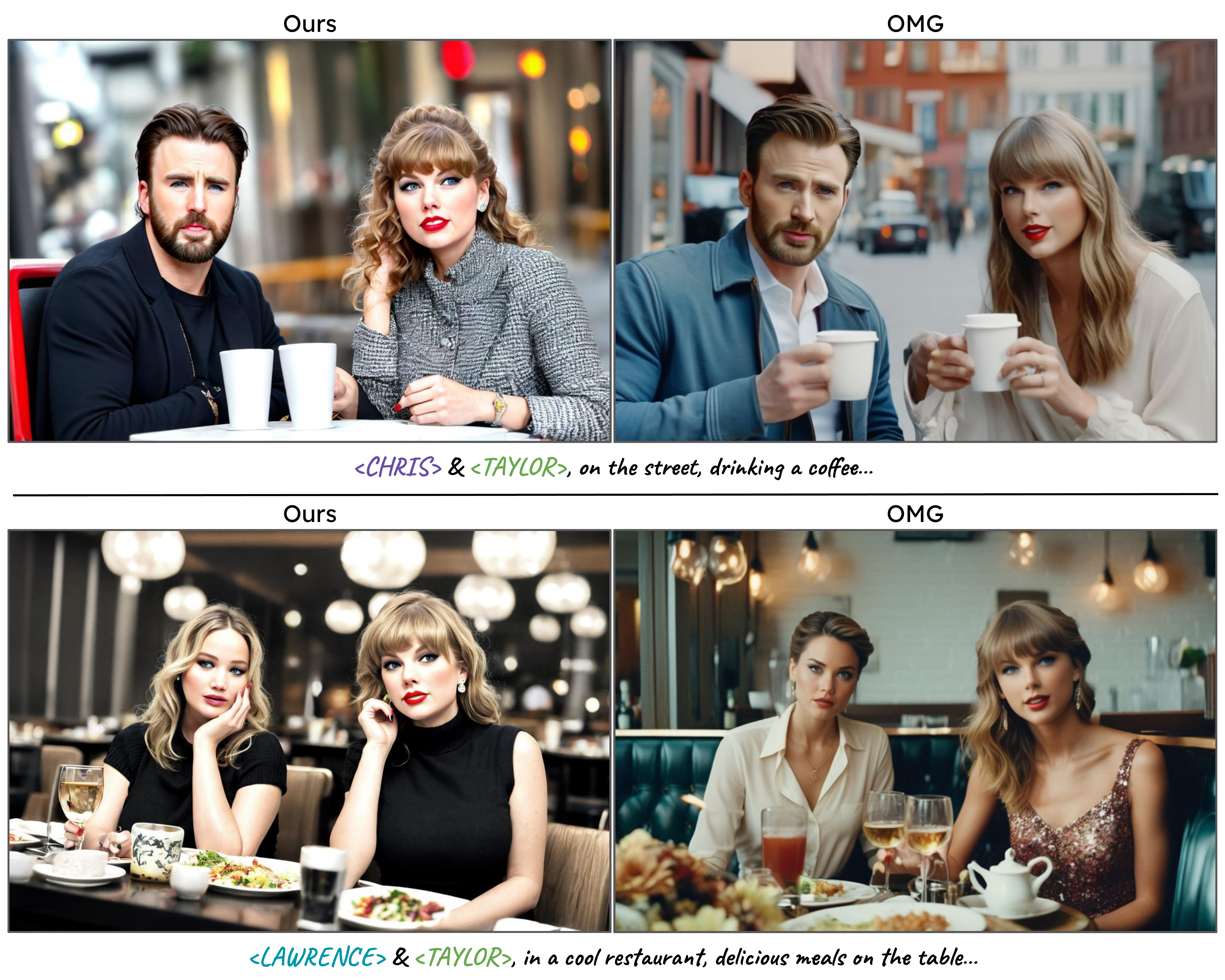}
    \caption{\textbf{Comparison between our method and OMG for generating multi-concept scenes.} OMG struggles with intermediate layout dependence and compositional errors, particularly with same-gender concepts, while our method achieves seamless and accurate results.}
    \label{fig:omg}
\end{figure*}

In contrast, our method bypasses the need for intermediate layouts, directly producing coherent and visually appealing compositions. As shown in ~\cref{fig:omg}, our approach excels in creating realistic and contextually aligned scenes, such as ``...on the street, drinking a coffee” and ``...in a cool restaurant, delicious meals on the table.” These examples highlight the superior fidelity and contextual understanding achieved by our method compared to OMG~\cite{kong2024omg}.

\subsection{More Qualitative Comparison}
This subsection provides additional qualitative results to highlight the strengths of our approach in generating multi-concept scenes, from 2 concepts to 6 concepts. Compared to existing methods such as Orthogonal Adaptation \cite{po2024orthogonal}, Mix-of-Show \cite{gu2023mix}, and $\mathcal{P}+$ \cite{voynov2023p+}, our method excels in producing coherent, contextually accurate, and visually appealing compositions, even in complex scenarios involving multiple concepts and intricate stylistic requirements.

\begin{figure*}
    \centering
    \includegraphics[width=1\linewidth]{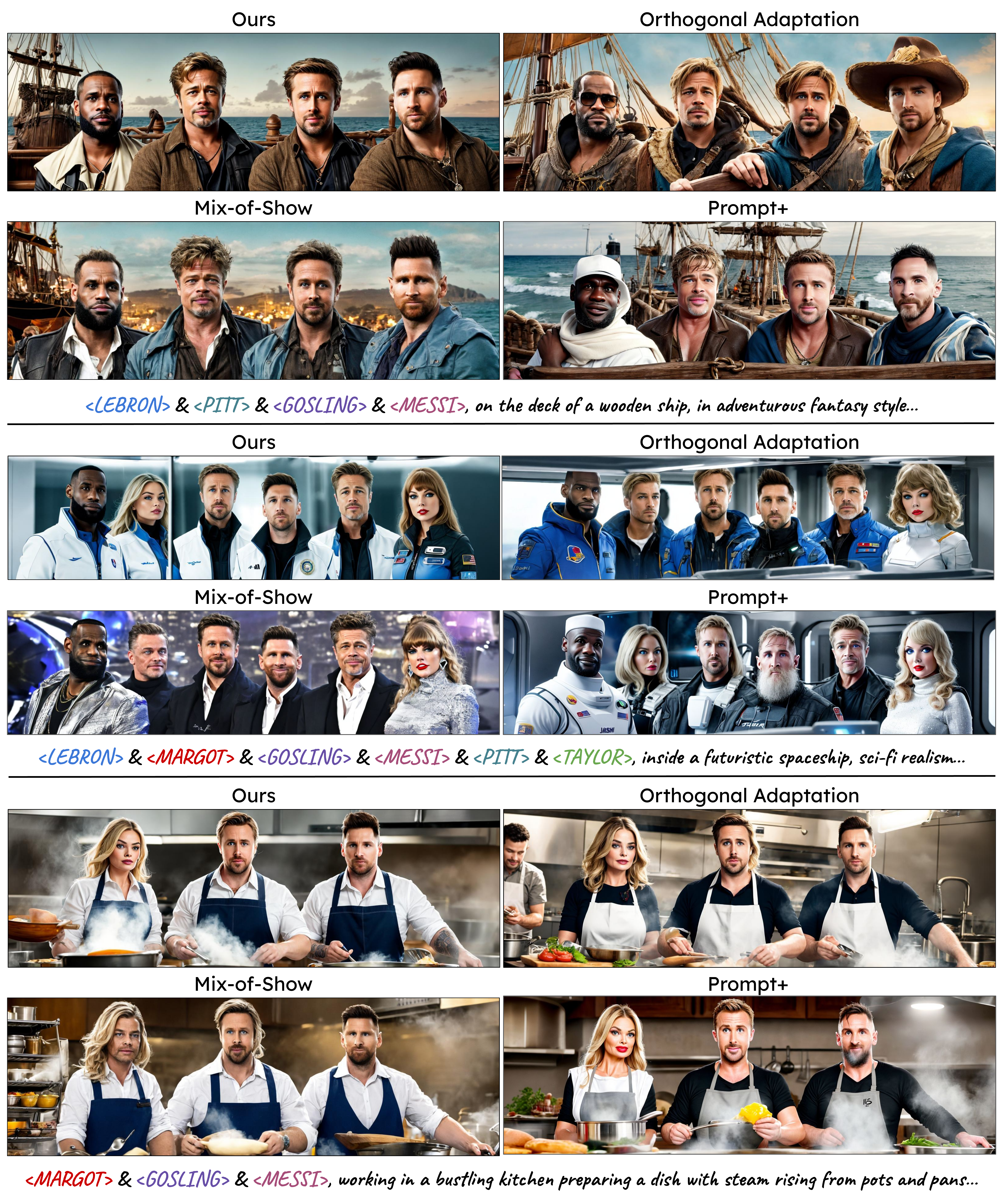}
    \caption{\textbf{Qualitative comparison of multi-concept scenes.} Our method effectively captures dynamic interactions and complex stylistic elements, as seen in examples such as bustling kitchens and futuristic spaceships. It surpasses Orthogonal Adaptation, Mix-of-Show and $\mathcal{P}+$ in coherence and realism.}
    \label{fig:comp1}
\end{figure*}

\Cref{fig:comp1} showcases examples such as ``...working in a bustling kitchen preparing a dish with steam rising from pots and pans.” Our method accurately captures the dynamic nature of the scene, ensuring proper interactions between concepts and retaining their distinct identities. In ``...inside a futuristic spaceship, sci-fi realism,” the futuristic aesthetics and intricate details are vividly rendered, demonstrating the superiority of our approach in handling complex compositions compared to baselines, which often introduce artifacts or fail to maintain consistency.

\begin{figure*}
    \centering
    \includegraphics[width=1\linewidth]{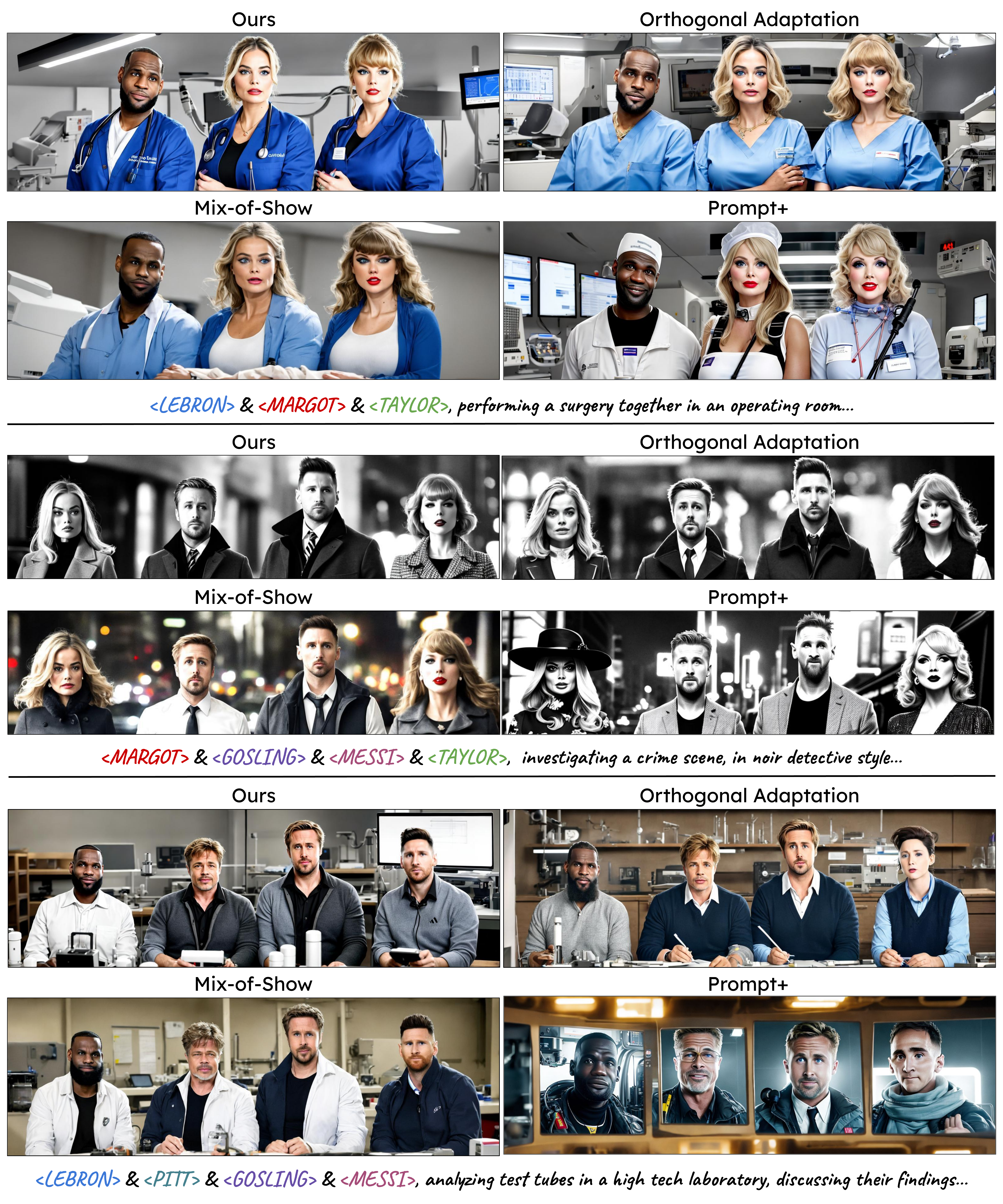}
    \caption{\textbf{Additional multi-concept image generation examples.} Our method demonstrates superior integration of concepts and themes in diverse scenarios, such as operating rooms and detective noir settings, while maintaining stylistic fidelity.}
    \label{fig:comp2}
\end{figure*}

\Cref{fig:comp2} further highlights the versatility of our method with scenes such as ``...performing a surgery together in an operating room.” Our model not only preserves the realism of the surgical environment but also ensures that all concepts are seamlessly integrated into the scene. In another example, ``...investigating a crime scene in noir detective style,” our method faithfully reproduces the intended stylistic elements while maintaining accurate subject interactions—a challenge for baseline methods that struggle to balance style and coherence.

\begin{figure*}
    \centering
    \includegraphics[width=1\linewidth]{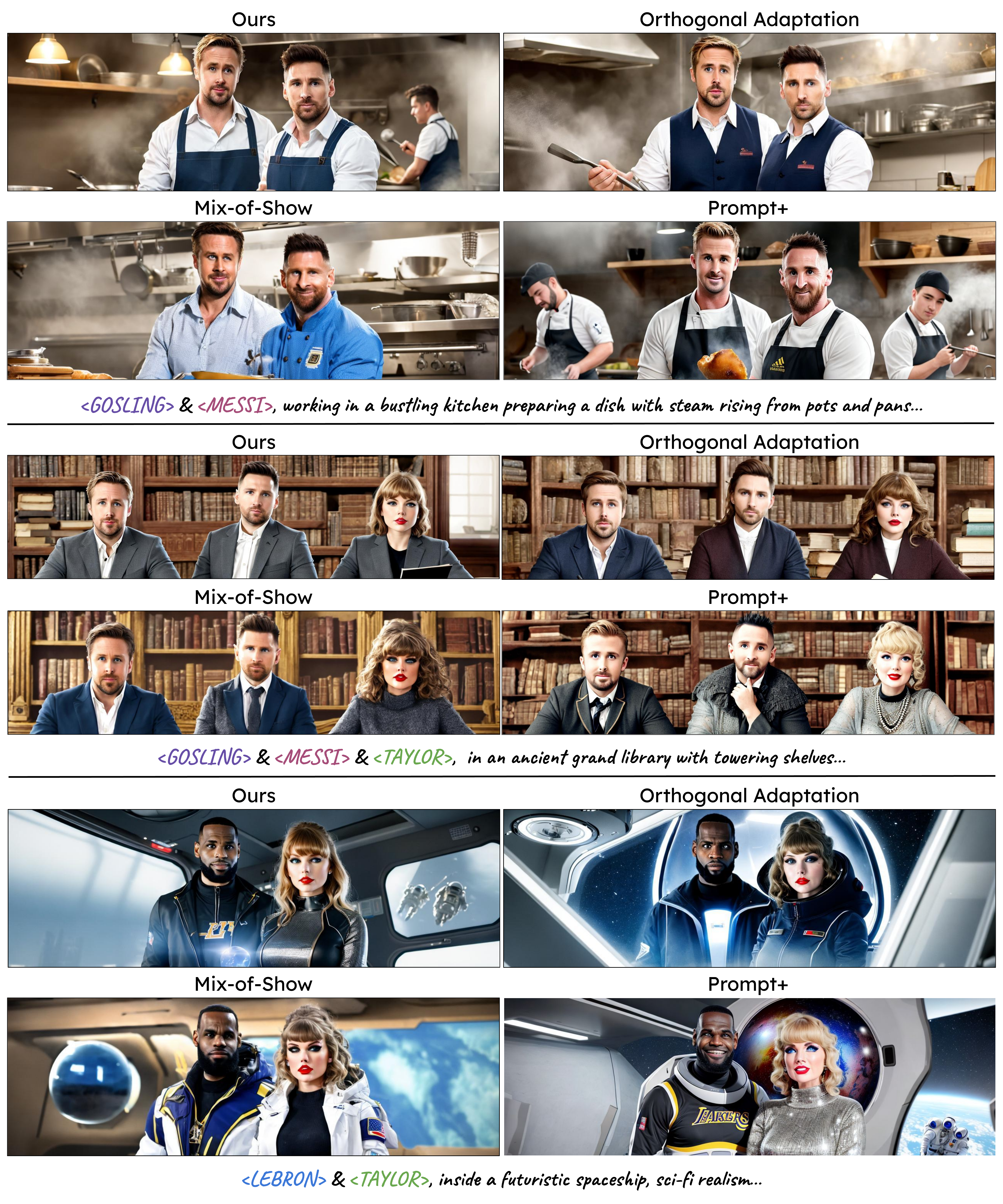}
    \caption{\textbf{Extended qualitative results for multi-concept image generation.} It showcases our method’s ability to generate intricate compositions, such as ancient libraries and sci-fi interiors. These results emphasize the robustness of our approach in maintaining style, subject integrity, and contextual relevance.}
    \label{fig:comp3}
\end{figure*}

Finally, \cref{fig:comp3} presents challenging scenarios like ``...in an ancient grand library with towering shelves.” Our method captures the details of the setting while ensuring the concepts interact naturally within the environment. In ``...inside a futuristic spaceship, sci-fi realism,” the vivid rendering of the scene’s futuristic details once again underscores the robustness of \model{} compared to baselines that exhibit inconsistencies in subject placement and stylistic alignment.

\section{User Study Details}
\begin{figure}[!htb]
    \centering
    \includegraphics[width=1\linewidth]{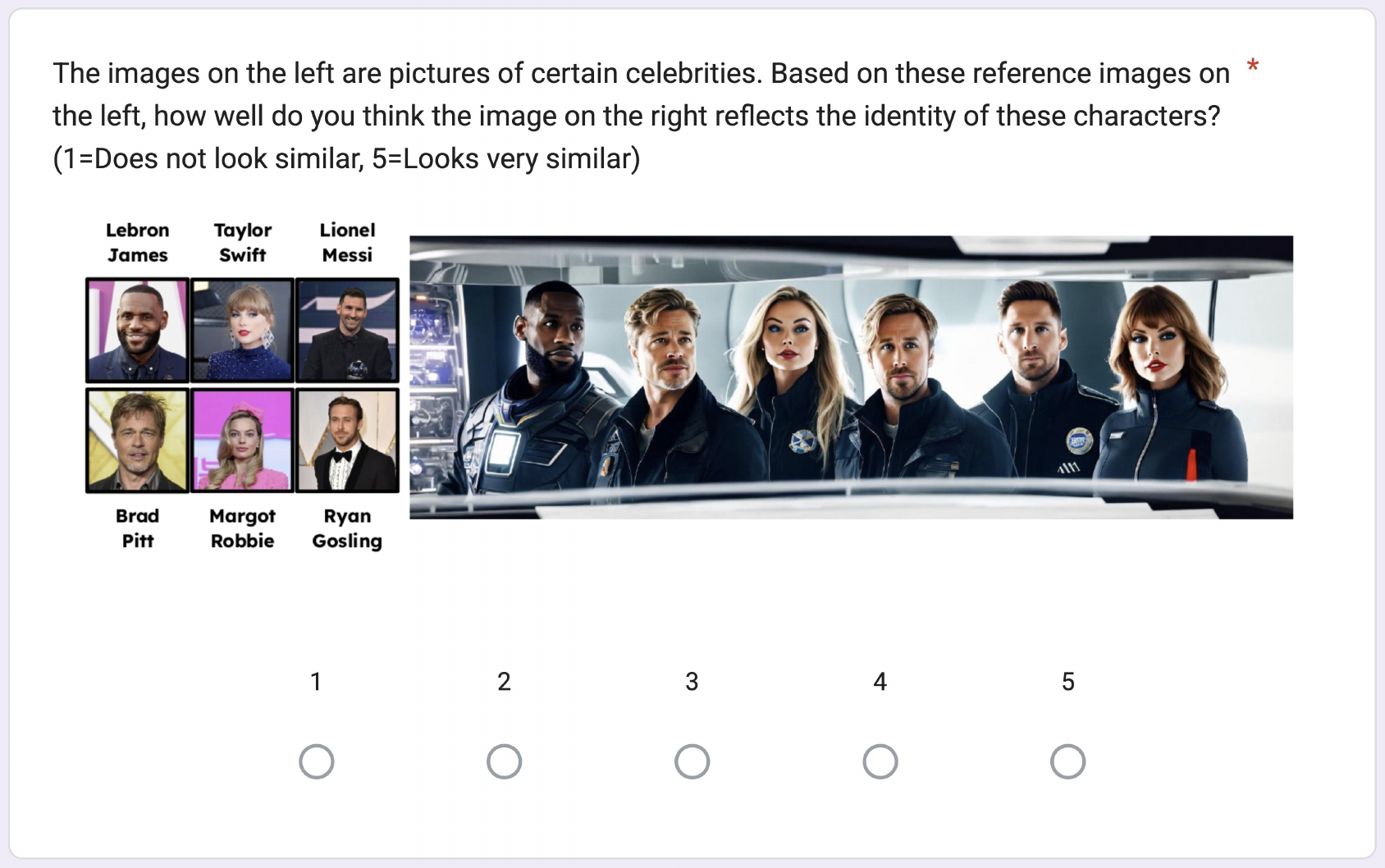}
    \caption{\textbf{User Study Interface.} Participants rated identity similarity between reference images and generated scenes, focusing on accuracy and realism.}
    \label{fig:user_study}
\end{figure}

We conducted a user study to evaluate identity preservation and composition quality in generated images. Participants were shown reference images alongside generated scenes (\cref{fig:user_study}) and asked to rate identity similarity on a scale of 1 (does not look similar) to 5 (looks very similar). 

\clearpage

\end{document}